# Tradeoffs for Space, Time, Data and Risk in Unsupervised Learning

**Mario Lucic**
ETH Zürich

**Mesrob I. Ohannessian**
University of California, San Diego

**Amin Karbasi**
Yale University

**Andreas Krause**
ETH Zürich

## Abstract

Faced with massive *data*, is it possible to trade off (statistical) *risk*, and (computational) *space* and *time*? This challenge lies at the heart of large-scale machine learning. Using $k$-means clustering as a prototypical unsupervised learning problem, we show how we can strategically summarize the data (control space) in order to trade off risk and time when data is generated by a probabilistic model. Our summarization is based on *coreset* constructions from computational geometry. We also develop an algorithm, TRAM, to navigate the space/time/data/risk tradeoff in practice. In particular, we show that for a fixed risk (or data size), as the data size increases (resp. risk increases) the running time of TRAM *decreases*. Our extensive experiments on real data sets demonstrate the existence and practical utility of such tradeoffs, not only for $k$-means but also for Gaussian Mixture Models.

## 1 INTRODUCTION

The computational and statistical performance of any learning algorithm for a given data set can be described in terms of three parameters: risk, running time, and space usage. The massive growth in datasets, coupled with limited resources in terms of time and space, raises new challenging questions on the accuracy of learning that can be achieved. At the heart of this challenge is to identify the relationships between *risk* $\varepsilon$, and the resources we have available, namely, *time* $t$, *space* $s$, and *data* $n$. Most of classical learning theory centers around the question of how risk scales with dataset (or sample) size: How much data $n$ is needed in order to achieve a certain level of risk $\varepsilon$ (i.e., what is the sample complexity of a given learning task)? In contrast, and from a practical point of view, increasing the data size is a source of computational complexity which typically translates into higher running time $t$. From this perspective, large data is considered a nuisance rather than a resource for achieving lower risk. As a result, most practical algorithms accumulate data until they exhaust either the time or space constraints and drop the data afterwards.

 

**Related Work.** An alternative direction is to investigate *computational and statistical tradeoffs*: using data as a computational resource when available beyond the sample complexity of the learning task. Pioneering this effort, Decatur et al. [2000] and Servedio [1999] showed tradeoffs in the realizable PAC learning model. Exploring these tradeoffs has gained much recent attention due to emerging problems in big data. For instance, Bottou and Bousquet [2008], Shalev-Shwartz and Srebro [2008] and Birnbaum and Shwartz [2012] showed the existence of such tradeoffs for learning linear classifiers as the data size increases. These tradeoffs are generally achieved by leveraging the fact that as we accumulate more data, the desired risk $\varepsilon$ becomes easier to reach, thus computationally cheaper but less accurate algorithms can be employed. This idea of *algorithmic weakening* was explored more systematically by Chandrasekaran and Jordan [2013] using convex relaxations.

**Our Contributions.** Existing approaches in computational and statistical tradeoffs consider only three of the four parameters: for a desired level of risk $\varepsilon$ they identify tradeoffs between running time $t$ and data size $n$. Our primary goal in this paper is to study how *summarization* (i.e., controlling space) can help navigate the tradeoff between time, data size and risk. In other words, we present a *weakening mechanism*, akin to Chandrasekaran and Jordan [2013], albeit in a different direction. Instead of weakening *learning algorithms*, we consider weakening the *data representation*. As more data becomes available, more representative elements can be extracted, without incurring much computational cost. Our approach is based on novel computational geometric techniques, called *coresets* (Agarwal et al. [2005]), where a small amount of most relevant data is extracted from the dataset, while performing the computation on this extracted data guarantees an approximate solution to the original problem. To the best of our knowledge, this paper is a first effort in introducing a methodological data-summarization approach for studying and navigating space/time/data/risk tradeoffs.

As a prototypical unsupervised learning problem, we focus on $k$-means clustering, also known as *vector quantization*, due to its simplicity and practical importance. In this problem, a set of $k$ centers is sought to minimize the expected (squared) distance between data points and the closest center. Finding the optimal centers is NP-hard, but good approximation algorithms are known, e.g., Lloyd's algorithm (Lloyd [1982]). We show how coreset constructions for $k$-means (Kanungo et al. [2002], Har-Peled and Mazumdar [2004], Agarwal et al. [2005], Feldman et al. [2007, 2013]) can be used to strategically summarize the data: in order to achieve a fixed precision, the running time can be made to *decrease* as the data set grows, by carefully controlling space usage. We also provide a practical algorithm TRAM that uses existing algorithms for solving $k$-means (e.g., Lloyd's algorithm, or $k$-means++) in order to realize this tradeoff in practice. We demonstrate the effectiveness of our summarization strategy on several synthetic and real data sets. We should highlight that $k$-means clustering is a *non-convex* problem, thus prior computational-statistical tradeoff strategies that heavily relied on convexity cannot be applied in this setting. While we focus on $k$-means, coresets are available for many other unsupervised learning tasks (Feldman et al. [2013]), and we believe that our approach can be applied much more generally. In particular, we empirically demonstrate how such tradeoffs can be achieved for Gaussian Mixture Models (GMMs).

## 2 THE STATISTICAL $k$-MEANS PROBLEM

Typically, $k$-means is viewed as a (combinatorial) optimization problem. We focus instead on the statistical variant. In particular, we assume that an underlying distribution generates i.i.d. samples, and we seek centers that generalize well. More formally, let $\mathbf{P}$ be an *unknown* distribution on $\mathbb{R}^d$ where we assume that it is supported on a ball of radius $B$ at the origin, i.e., for $X \sim \mathbf{P}$ we have $\mathbf{P}(\|X\|_2 \leq B) = 1$ (this assumption can be relaxed under other regularity conditions, see, for example Telgarsky and Dasgupta [2013]). In $k$-means clustering, any data point $x \in \mathbb{R}^d$ is associated with the closest among a set of $k$ *centers* $c = \{c_1, \cdots, c_k\}$, where $c_i \in \mathbb{R}^d$. We judge the quality of this association by a *risk* defined as

$$R(c) = \mathbf{E}_{X\sim\mathbf{P}}[\mathrm{d}^2(c,X)]$$

between $c$ and a sample $X$ from $\mathbf{P}$, where $\mathrm{d}^2(c,X) = \min_{i=1}^k ||c_i - X||_2^2$. Let $\mathscr{C}$ be the set of all $k$ centers in the ball of radius $B$ at the origin. The *optimal centers* are those that minimize this risk:

$$c^\star = \arg\min_{c\in\mathscr{C}} R(c).$$

The solution to this minimization may not be unique, but for the ease of presentation we assume it is. We further make the realistic assumption that $\underline{R} := R(c^\star) > 0$ which is satisfied for any distribution supported on more than $k$ points. Since $\mathbf{P}$ is unknown, we seek centers for a dataset of $n$ samples $X_1, \ldots, X_n$ drawn i.i.d. from $\mathbf{P}$. Any choice of a sequence of functions $\tilde{c}_n$, from $\mathbb{R}^{d\times n} \to \mathbb{R}^{d\times k}$ is called a $k$-means *procedure*. Out of all such choices, of particular importance is the one that minimizes the *empirical risk*, to obtain the *empirically optimal centers*:

$$R_n(c) = \frac{1}{n}\sum_{i=1}^{n} \mathrm{d}^2(c,X_i), \quad \hat{c}_n = \arg\min_{c\in\mathscr{C}} R_n(c). \qquad (1)$$

The properties of the empirically optimal centers have been extensively studied in the literature ([Kanungo et al., 2002, Ben-David, 2007]). In particular, finding empirically optimal centers is a daunting task and often approximate procedures are used. Of particular interest to us is a class of algorithms (Kanungo et al. [2002], Har-Peled and Mazumdar [2004], Agarwal et al. [2005], Feldman et al. [2007, 2013]) that solve the $k$-means problem by first summarizing the data and then finding the centers on the summarized data. This decoupling principle allows these algorithms to invest most of their running time only on a small set of points and, at the same time, to save space.

## 3 DATA SUMMARIZATION

Data summarization refers to a procedure that takes a data set of size $n$ and replaces it with a smaller set of size $s_{\mathsf{proc}}$, which (approximately) suffices for solving the learning task at hand. This summarization may simply be a truncation without any consideration to the inherent structure of the data (a simple method that is often practiced), or it may be a combination of truncation and strategic sampling that adapts to structure in the data. We denote the truncation size by $m_{\mathsf{proc}}$. One of the main advantages of having summarized data, apart from saving space, is the substantial reduction in running time. For this reason, truncation must be allowed, as otherwise the running time of *any* learning algorithm would grow with the data size. We now formally present these two strategies.

**Uniform Subsampling** This is the simplest form of data summarization: start with a data set of size $n$, preserve only the first $s_{\mathsf{subs}} \leq n$ points, and then solve the learning problem by minimizing the empirical risk. In the $k$-means problem, this amounts to $\tilde{c}_{\mathsf{subs}} = \arg\min_{c\in\mathscr{C}} R_{s_{\mathsf{subs}}}(c)$ where $R_{s_{\mathsf{subs}}}(c) = \frac{1}{s_{\mathsf{subs}}}\sum_{i=1}^{s_{\mathsf{subs}}} \mathrm{d}^2(c,X_i)$. For the uniform subsampler the summarization and truncation sizes are identical, $s_{\mathsf{subs}} = m_{\mathsf{subs}}$. Larger values of $s_{\mathsf{subs}}$ promote lower statistical risk but are more expensive to compute. Conversely, computation on a smaller set may be fast but results in higher risk. The uniform subsampler may tune $s_{\mathsf{subs}}$ to balance risk with running time.

**Strategic Sampling** *Coresets* are data summaries that are constructed via adaptive sampling, in the spirit of importance sampling. As with the uniform subsampler, we start with data of size $n$, then truncate it to $m_{\mathsf{core}}$ points. Now, instead of using the truncation as is, we perform strategic sampling to propose a set of $s_{\text{core}}$ representative points $(Y_j)_{j=1,\cdots,s_{\text{core}}}$, each associated with a non-negative weight $w_j$, and we solve the learning problem not on the empirical risk, but on a *weighted* variant. In the $k$-means problem, this amounts to $\tilde{c}_{\text{core}} = \arg\min_{c\in\mathscr{C}} R^w_{s_{\text{core}}}(c)$ where $R^w_{s_{\text{core}}}(c) = \sum_{j=1}^{s_{\text{core}}} w_j \mathrm{d}^2(c, Y_j)$. Coresets strive to be a more faithful/concise representation of the data than uniform samples. Naturally, their properties depend on how the strategic sampling is performed. The hallmark property of coresets is their ability to approximate the empirical risk, defined in (1), optimized over the starting $m_{\text{core}}$ data points.

**Definition 1.** *A coreset construction is a* $(1+\eta)$*-approximation, with* $\eta$ *a function of the coreset size* $s_{\text{core}}$*, if the centers* $\tilde{c}_{\text{core}}$ *satisfy* $R_{m_{\text{core}}}(\tilde{c}_{\text{core}}) \leq (1+\eta(s_{\text{core}}))R_{m_{\text{core}}}(\hat{c}_{m_{\text{core}}})$*.* [1]

A coreset procedure could start out with a moderately larger truncation $m_{\text{core}} > m_{\text{subs}}$, and yet produce a representation that is significantly smaller $s_{\text{core}} \ll s_{\text{subs}}$, all while maintaining a comparable risk. Note again that without performing truncation, the running time of finding a coreset of size $s_{\text{core}}$ using the whole dataset grows with the data size. A number of efficient $(1+\eta)$-coreset constructions for $k$-means are known, as reviewed in Section 5.2. We study a particularly practical variant in Section 7. Additionally, it is worth noting that coresets have the advantage of admitting *streaming* and *parallel* constructions (Har-Peled and Mazumdar [2004], Balcan et al. [2013]), which makes them particularly suited for massive datasets.

## 4 SPACE-TIME-DATA-RISK TRADEOFF

Our goal now is to give a precise definition of tradeoffs: how data summarization may lead to trading off representation space, running time, data size, and statistical risk. Let $\tilde{c}_{\text{proc}}(n, m_{\text{proc}}, s_{\text{proc}})$, or $\tilde{c}_{\text{proc}}$ for short, denote a $k$-means procedure based on data summarization, such as uniform subsampling or coreset summarization. Recall that such a procedure starts with $n$ data points, truncates them to $m_{\text{proc}}$ points, summarizes these to $s_{\text{proc}}$ (possibly weighted) representative points, and optimizes the (possibly weighted) empirical risk to obtain the set of centers $\tilde{c}_{\text{proc}}$. The *running time*, which we denote by $t_{\text{proc}}$, may be further decomposed into: summarization time $t^{\text{sum}}_{\text{proc}}$ and the time $t_{\text{solver}}$ for empirical risk optimization. The former depends on the particular procedure, but the latter can be a generic solver across procedures. We assume that the act of truncation (for both the uniform subsampler and the coreset procedure) has no computational cost. The *statistical risk* of the procedure, which we denote by $R_{\text{proc}}$, is the expected risk, where the expectation is taken with respect to the sample. That is, $R_{\text{proc}} := \mathbf{E}[R(\tilde{c}_{\text{proc}})]$. We can decompose it as follows:

$$\begin{aligned} R_{\text{proc}} \;\leq\;& \underbrace{R(c^\star)}_{\varepsilon_{\text{model}}} + \underbrace{\mathbf{E}[R(\hat{c}_{m_{\text{proc}}})] - R(c^\star)}_{\varepsilon_{\text{est}}} \\ &+ \underbrace{|\mathbf{E}[R(\tilde{c}_{\text{proc}})] - \mathbf{E}[R(\hat{c}_{m_{\text{proc}}})]|}_{\varepsilon_{\text{sum}}}, \end{aligned} \tag{2}$$

where $\varepsilon_{\text{model}}$, $\varepsilon_{\text{est}}$, and $\varepsilon_{\text{sum}}$ are the *modeling*, *estimation*, and *summarization* errors, respectively. The modeling error is the best risk achieved by any $k$ centers (limitation of the model). The estimation error is incurred due to using the *empirically* optimal centers (limitation of estimating from data). Lastly, we have the error of *approximate* data summarization. For coresets it depends on $\eta$ (cf. Proposition 4).

**How to trade off** The four dimensions *space, time, data, and risk* put forth in this paper can now be represented by the four parameters $(s_{\text{proc}}, t_{\text{proc}}, m_{\text{proc}}, R_{\text{proc}})$. We can obtain a variety of tradeoffs by constraining some dimensions and optimizing others. Of course, not all $(s,t,m,R)$-tuples are attainable: for instance, classical sample complexity bounds constrain what risks are attainable at what data sizes. We call a subset of the dimensions *feasible* for a procedure, if there exist values of the others that lead to attainable tuples. By exploring the feasible landscape, one can harness various trends. For example, based on the risk decomposition stated above, as we decrease $s_{\text{proc}}$, the risk $R_{\text{proc}}$ increases due to the increase in $\varepsilon_{\text{sum}}$. In contrast, solving the optimization becomes computationally cheaper with smaller $s_{\text{proc}}$. These interactions, illustrated schematically in Figure 1b give rise to various tradeoffs. Some of these are listed in Figure 1a.

In this paper, we are mainly interested in *(a) data-time tradeoffs*: for $R_{\text{proc}}$ fixed below some $\varepsilon_{\text{total}}$, can $t_{\text{proc}}$ decrease as $n$ increases? and *(b) risk-time tradeoffs*: for some fixed $n$, can $t_{\text{proc}}$ decrease as $R_{\text{proc}}$ increases? These two tradeoffs are listed respectively in the first and second rows of the table in Figure 1a. Data summarization gives us a natural framework to answer those questions: we could achieve such gains by optimizing summarization space $s_{\text{proc}}$. This captures the weakening-through-data-summarization mechanism that we advocate in this paper. Formally, given a data size $n$ and risk $\varepsilon_{\text{total}}$, the *optimal running time* function is:

$$\begin{aligned} t^\star_{\text{proc}}(n, \varepsilon_{\text{total}}) = \min\nolimits_{m_{\text{proc}}, s_{\text{proc}}} t_{\text{proc}}(n, m_{\text{proc}}, s_{\text{proc}}), \\ \text{s.t. } R_{\text{proc}}(m_{\text{proc}}, s_{\text{proc}}) \leq \varepsilon_{\text{total}}, m_{\text{proc}} \leq n. \end{aligned} \tag{3}$$

Observe that for fixed $\varepsilon_{\text{total}}$ and as $n$ varies, the optimal running time $t^\star_{\text{proc}}$ is non-increasing in $n$ by construction. Similarly, for fixed $n$ and as $\varepsilon_{\text{total}}$ varies, the optimal running time $t^\star_{\text{proc}}$ is non-increasing in $\varepsilon_{\text{total}}$.

**Definition 2.** *We say that a k-means procedure offers a (non-trivial)* data-time tradeoff *if, for a given desired total risk* $\varepsilon_{\text{total}}$*, the running time* $t^\star_{\text{proc}}(\cdot, \varepsilon_{\text{total}})$ *is decreasing*

[1] Coresets conventionally require approximating the risk at *all* $c$: for $\varepsilon \in (0,1)$, $\forall c \in \mathscr{C}$, $|R^w_{s_{\text{core}}}(c)/R_m(c) - 1| \leq \varepsilon$. This implies a $(1+\eta)$-approximation with $\eta = 2\varepsilon/(1-\varepsilon)$.

| Tradeoff | Space | Time | Data | Risk |
|---|---|---|---|---|
| Data-Time | Tune | Objective | Vary | Fixed |
| Risk-Time | Tune | Objective | Fixed | Vary |
| Space-Risk | Vary | Tune | Fixed | Objective |
| Data-Risk | Tune | Fixed | Vary | Objective |
| Space-Time | Vary | Objective | Tune | Fixed |

(a) Tradeoffs

| Cost | k$\nearrow$ | n$\nearrow$ | s$\nearrow$ |
|---|---|---|---|
| $\varepsilon_{\text{modelling}}$ | $\searrow$ | | |
| $\varepsilon_{\text{estimation}}$ | $\nearrow$ | $\searrow$ | |
| $\varepsilon_{\text{summarization}}$ | $\nearrow$ | $\rightarrow$ | $\searrow$ |
| $t$ | $\nearrow$ | $\nearrow$ | $\nearrow$ |

(b) Trends

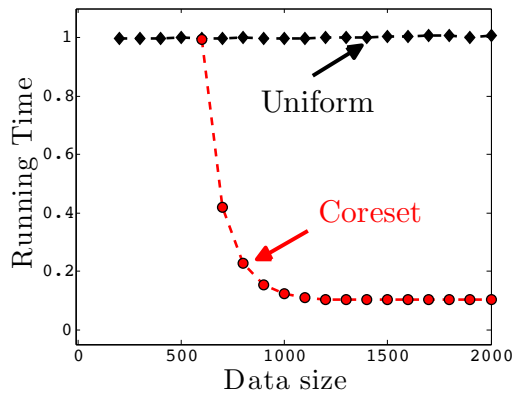

(c) Data-Time

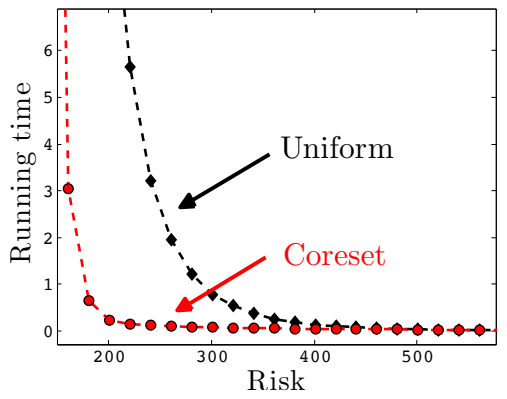

(d) Risk-Time

Figure 1: (a) Examples of Space-Time-Data-Risk-Tradeoffs, each realized by trading off two parameters (green and gray), by constraining (red) and tuning (blue) the remaining ones. (b) Effect of increasing $k$, $n$ and $s$ on the various errors and running time $t$. (c) Coreset (red) data-time tradeoffs versus subsampler (black). The plots represent best running time for fixed risk tolerance when varying the data size, as predicted by our theory (Section 5). (d) Risk-time tradeoff, i.e., best achievable running time for fixed data size when varying the allowed risk. [Time units normalized to the median subsampler time.]

*for some range of $n$. We say that the procedure offers a (non-trivial)* risk-time tradeoff *if, for a given data size $n$, $t^\star_{\text{proc}}(n,\cdot)$ is decreasing for some range of $\varepsilon_{\text{total}}$. In other words, these tradeoffs correspond to (non-flat)* Pareto optimal frontiers *of $t^\star_{\text{proc}}$, as either of the arguments is fixed.*

Tradeoffs divide the landscape into various operation regimes. For data-time tradeoffs, before $n$ reaches the feasible range for $\varepsilon_{\text{total}}$, we are in a "data-bounded" regime (cf., Shalev-Shwartz and Srebro [2008]). We cannot get the desired risk $\varepsilon_{\text{total}}$, and have to invest all of the data and computation to driving the risk as low as possible. On the other extreme, very large data sizes are bound to lead to a point where more data can safely be discarded with no further impact on risk and computation time. This is the "data-laden" regime. In our framework, it means that in the data-laden regime $t^\star_{\text{proc}}(\cdot,\varepsilon_{\text{total}})$ flattens. Lastly, there is an "intermediate regime" where all of the available data is used, but there is maneuvering room to drive the computation time down or in other words $t^\star_{\text{proc}}(\cdot,\varepsilon_{\text{total}})$ decreases. A lot of the subtlety of the tradeoffs happens in this regime. We see this phenomenon manifest itself both analytically, in Section 5, and experimentally, in Section 7.

**Extensions** Our methodology is formalized for the $k$-means problem, but the framework is much richer. For example, spectral clustering methods that can be mapped to $k$-means are bound to profit directly from our results. A concrete extension consists of Gaussian Mixture Models, by using the negative log-likelihood as the risk and coreset construction by Feldman et al. [2011]. We do not formalize this, but we demonstrate it experimentally in Section 7.

## 5 ANALYSIS

We have thus far motivated and laid out a clear paradigm of tradeoffs via data summarization. But are such tradeoffs even possible? In this section, we show that the answer is *yes*. To keep our exposition concise, we focus in particular on showing that nontrivial data-time tradeoffs (Definition 2) do indeed exist. For this we need to characterize $t^\star_{\text{proc}}(n,\varepsilon_{\text{total}})$ as $n$ varies, for a fixed desired risk level $\varepsilon_{\text{total}}$.

For the uniform subsampler the data-time tradeoff is necessarily trivial. To see this, let $n_{\text{f}}(\varepsilon_{\text{total}})$ be the smallest data size $n$ when $\varepsilon_{\text{total}}$ becomes feasible. Then for all $n \geq n_{\text{f}}$ we have $\varepsilon_{\text{model}} + \varepsilon_{\text{est}}(n) \leq \varepsilon_{\text{total}}$ but the uniform subsampler has no incentive to use more than $m_{\text{subs}} = n_{\text{f}}$ samples, since otherwise its running time would be greater (for unneeded risk reduction). This means that $t^\star_{\text{subs}}(\cdot,\varepsilon_{\text{total}})$ is undefined for $n < n_{\text{f}}(\varepsilon_{\text{total}})$, and is flat beyond that. In the language of Section 4, the uniform subsampler switches abruptly from the "data-bounded" to the "data-laden" regime.

The more interesting question is thus: Can coreset procedures give non-trivial data-time tradeoffs that improve on the uniform subsampler? In particular, can we observe an "intermediate regime" where $t^\star_{\text{core}}(\cdot,\varepsilon_{\text{total}})$ curves down, before reaching the data-laden regime? Our main result answers these questions in the affirmative. Informally, we have the following.

**Main Result** (Existence of Tradeoffs)**.** *Let the following conditions hold for a coreset procedure:*

*(a) The summarization is time-efficient (its running time is negligible relative to that of the solver).*

*(b) The summarization is sample-efficient (the approximation factor vs. summarization size decays faster than the estimation error vs. sample size).*

*(c) The estimation error decays fast ($\sim$ power law).*

*(d) The solver is slow (at least super-linear).*

*Then, for small enough risks, the procedure admits a non-trivial tradeoff, and its optimal running time dominates (is less than) that of the uniform subsampler for large enough data sizes. Moreover, existing bounds and coreset constructions do satisfy these conditions.*

In what follows, we proceed to formalize this result. In Section 5.1 we give the sufficient conditions and in Section 5.2 we affirm that these conditions are satisfied in practice, by giving existing risk bounds and coreset constructions. We also provide some numerical illustrations of tradeoffs using these bounds. In Section 7 we demonstrate these tradeoffs experimentally.

### 5.1 Sufficient Conditions for Tradeoffs

Recall first some notation from Section 4. When a coreset summarization procedure has a total risk ($R_{\text{core}}$), it can be decomposed into modeling ($\varepsilon_{\text{model}}$), estimation ($\varepsilon_{\text{est}}$), and summarization errors ($\varepsilon_{\text{sum}}$). The latter depends on the coreset approximation that results from a choice of a given summarization size ($\eta(s_{\text{core}})$) (Proposition 4 makes this precise). The total running time of the procedure $t_{\text{proc}}$ can be decomposed into summarization time ($t_{\text{proc}}^{\text{sum}}$) and empirical risk minimization time ($t_{\text{solver}}$). The latter is attributed to a generic solver, and it depends only on the size ($s_{\text{proc}}$) of its input. For the former, we add some further notation due to "bicriteria"-type coreset constructions (Feldman and Langberg [2011]), where the summarization stage itself is decoupled into two: initialization, taking time $t_{\text{core}}^{\text{init}}(m_{\text{core}})$ that depends only on the (truncated) data size, followed by adaptive sampling, with time $t_{\text{core}}^{\text{samp}}(s_{\text{core}})$ that depends only on the coreset summarization size. We are now ready to formally state our main result's conditions.

**Theorem 1.** *Let $t_{\text{solver}}(\cdot)$, $t_{\text{core}}^{\text{init}}(\cdot)$, $t_{\text{core}}^{\text{samp}}(\cdot)$ be increasing, and $\varepsilon_{\text{est}}(\cdot)$ and $\eta(\cdot)$ be decreasing functions of their arguments. Let the setting of the coreset procedure be such that the following are satisfied:*

*(a) $t_{\text{core}}^{\text{init}}(\cdot)$ is linear and $t_{\text{core}}^{\text{samp}}(x) = o(t_{\text{solver}}(x))$,*

*(b) $\exists a, b > 0$ such that for large enough $x$, $2\eta(x) \leq (1/\varepsilon_{\text{model}} - a)\,\varepsilon_{\text{est}}((1+b)x)$.*

*(c) $\forall L(x) \to \infty$, no matter how slowly, $\frac{\varepsilon_{\text{est}}(xL(x))}{\varepsilon_{\text{est}}(x)} \to 0$, as $x \to \infty$,*

*(d) $t_{\text{solver}}(\cdot)$ is bounded from below by a convex super-linear function, i.e. $\frac{t_{\text{solver}}(x)}{x} \to \infty$, as $x \to \infty$,*

*Then there exists a small enough risk $\varepsilon_0$, such that for all desired risks $\varepsilon_{\text{total}} \leq \varepsilon_0$, there exists a large enough sample size $n_0$, beyond which for all $n > n_0$ we have $t_{\text{core}}^{\star}(n, \varepsilon_{\text{total}}) < t_{\text{subs}}^{\star}(n, \varepsilon_{\text{total}})$.*

Since the coreset procedure cannot be faster than the subsampler at a sample size at the threshold of feasibility, the theorem implies that for all $\varepsilon_{\text{total}} \leq \varepsilon_0$ the coreset procedure achieves a non-trivial tradeoff with an "intermediate regime", eventually dominating the uniform subsampler for large enough sample sizes.

Condition (a) asks for the solver's running time to overshadow that of summarization (how could one benefit from summarization otherwise?). Slower solvers can only "help" satisfy this condition. Condition (b) is more subtle, though it can be understood as follows: if larger summaries do not drive the summarization error down as fast as larger sample sizes drive the estimation error down, then summarization loses its competitive advantage against truncation. As for Conditions (c) and (d), they are primarily used in a technical context, to balance asymptotic expressions. As we outline in Section 5.2, these conditions are natural behaviors for the estimation error and solver respectively.

*Proof sketch of Theorem 1.* To prove this theorem, it suffices to show that for a large enough sample size $x$ we can find a (possibly suboptimal) coreset size $s$ such that the resulting procedure has $\varepsilon_{\text{total}} = \varepsilon_{\text{model}} + \varepsilon_{\text{est}}(x)$ while its running time is less than $t_{\text{solver}}(x)$. This is because $x$ and $t_{\text{solver}}(x)$ represent respectively the feasibility threshold and the optimal running time of the uniform subsampler that achieves a risk of $\varepsilon_{\text{total}}$. To maintain a risk of $\varepsilon_{\text{total}}$, the corset procedure needs to choose an appropriate truncation size $m$ slightly larger than $x$, and the result would only hold for $n \geq m$, thus allowing enough samples for summarization.

We make a simple choice, $s = t_{\text{solver}}^{-1}((1-2\delta)t_{\text{solver}}(x))$ for some $\delta > 0$, ignoring rounding. This implies a performance gap $t_{\text{solver}}(x) - t_{\text{solver}}(s)$ of $2\delta t_{\text{solver}}(x)$ within which we can maneuver. Then Condition (a) implies that for large enough $x$ the sampling stage will occupy less than $\delta t(x)$ of this gap. On the other hand, the initialization stage depends linearly on the resulting $m$. Condition (b) then intervenes to show that the impact of this stage remains also within another $\delta t(x)$, thus establishing the theorem. This, however, also requires $x$ to be large enough to align with the constants of Condition (b), and for that we invoke Conditions (c) and (d). The details can be found in the supplements. □

### 5.2 Existence of Tradeoffs

We now affirm that the conditions of Theorem 1 are met by existing constructions.

**Proposition 1.** *Under known risk bounds (Propositions 2 and 3) and coreset constructions (Feldman and Langberg [2011]), and when using a super-linear polynomial-time or slower solver, the conditions of Theorem 1 are satisfied.*

We can illustrate this result visually via simulations: we perform numerical optimization using the risk, running time, and summarization bounds given in this section. The details can be found in the supplements. We plot a representative data-time tradeoff of both the subsampler (in black) and the coreset procedure (in red) in Figure 1c. Note that the coreset procedure dominates. The same type of numerical optimization can be done to obtain other tradeoffs: we plot the risk-time tradeoff of the same problem in Figure 1d. As Theorem 1 predicts, the coreset dominates primarily for smaller values of the risk. The proof of Proposition 1, also in the supplments, is a direct verification of the conditions of Theorem 1. We give here an account of the invoked bounds and coreset construction.

**Risk Bounds** The following bounds characterize the risks in terms of the parameters of the problem: the dimension $d$, radius $B$, and number of clusters $k$. Note that the modeling error does not depend on the procedure, the estimation error only depends on the procedure through the truncation size $m_{\text{proc}}$, and the summarization errors depend more closely on the specifics of the summarization. The following bound on the modeling error is minimax up to constants (Graf and Luschgy [2000]).

**Proposition 2** (Modeling Error)**.** *The modeling error satisfies* $\varepsilon_{\mathsf{model}} \leq \frac{B^2 d}{k^{2/d}}$.

The estimation error has been extensively studied in statistics. We have the following (Antos et al. [2005]):

**Proposition 3** (Estimation Error)**.** *The estimation error satisfies* $\varepsilon_{\mathsf{est}} \leq \overline{\sigma} B^2 \frac{\sqrt{kd}}{\sqrt{m_{\mathsf{proc}}}}$*, for some* $\overline{\sigma} > 0$*. Furthermore, we have a lower bound: there exists* $\underline{\sigma} > 0$ *such that whenever* $k \geq 3$*, we may find* $\mathbf{P}$ *for which for large enough* $m_{\mathsf{proc}}$ *we have:* $\varepsilon_{\mathsf{est}} \geq \underline{\sigma} B^2 \frac{\sqrt{k^{1-4/d}}}{\sqrt{m_{\mathsf{proc}}}}$.

The summarization error depends on the particular summarization procedure. For uniform subsampling, since $\tilde{c}_{\mathsf{subs}} = \hat{c}_{m_{\mathsf{subs}}}$, it is trivially zero (cf. Equation (2)). For a coreset procedure, it depends on the coreset size or equivalently the approximation factor $\eta$.

**Proposition 4** (Summarization Error)**.** *Given a* $(1+\eta)$*-approximation coreset, when* $\eta(s_{\mathsf{core}}) \geq \eta_0 > 0$*, then* $\varepsilon_{\mathsf{sum}} < 2(\varepsilon_{\mathsf{model}} + \varepsilon_{\mathsf{est}})\eta(s_{\mathsf{core}})$ *for large enough m.*

Propositions 2 and 3 are restatements. On the other hand, Proposition 4 is new. The proof relies on uniform concentration, and is detailed in the supplements.

**Running Time Bounds** Solving for the exact empirically optimal centers is NP-hard, with the running time of known exact algorithms being $t_{\mathsf{solver}}(s) = \Omega(s^{kd})$ (cf., Inaba et al. [1994]). There are various popular heuristics, including Lloyd's ("the *k*-means") algorithm, and on typical inputs these have polynomial running times $t_{\mathsf{solver}}(s) = \Omega(\text{poly}(k)\text{poly}(d)\text{poly}(s))$. Under further conditions they can be exact (Meyerson et al. [2004]). Even these optimistic polynomial running times are sufficient for us.

The uniform subsampler performs no summarization beyond truncation, $s_{\mathsf{subs}} = m_{\mathsf{subs}}$. Thus $t^{\mathsf{sum}}_{\mathsf{subs}} = 0$, and:

$$t_{\mathsf{subs}}(n, m_{\mathsf{subs}}, s_{\mathsf{subs}}) = t_{\mathsf{solver}}(s_{\mathsf{subs}}).$$

For coresets, we use the above-mentioned "bicriteria" construction by Feldman and Langberg [2011]. We have:

$$t_{\mathsf{core}}(n, m_{\mathsf{core}}, s_{\mathsf{core}}) = t_{\mathsf{solver}}(s_{\mathsf{core}}) + t^{\mathsf{init}}_{\mathsf{core}}(m_{\mathsf{core}}) + t^{\mathsf{samp}}_{\mathsf{core}}(s_{\mathsf{core}}). \quad (4)$$

Like the risks, these initialization and sampling times depend on the various parameters of the problem, and in particular the dimension $d$ and the number of clusters $k$. In many constructions, these are *linear* functions of their arguments. In particular, the coreset construction of Feldman and Langberg [2011] is a $(1+\eta)$-approximation with $t^{\mathsf{init}}_{\mathsf{core}}(m_{\mathsf{core}}) = O(dkm_{\mathsf{core}})$ and $t^{\mathsf{samp}}_{\mathsf{core}}(s_{\mathsf{core}}) = O(s_{\mathsf{core}})$.

**Coreset Approximation** The last component of Proposition 1, needed to fully characterize a coreset approximation, is the functional relationship between the approximation factor $\eta$ and the coreset size $s_{\mathsf{core}}$. In particular, we note that Feldman and Langberg [2011] gives a $(1+\eta)$-approximation with a coreset of size $s_{\mathsf{core}} = O(dk(2+\eta)^2/\eta^2)$ (see[1] for reference). We may thus write $\eta(s_{\mathsf{core}}) = O\left(\sqrt{dk}/(\sqrt{s_{\mathsf{core}}} - \sqrt{dk})\right)$.

# 6 DATA-DRIVEN TRADEOFF NAVIGATION

So far we demonstrated tradeoffs in *k*-means by considering analytical models. In practice, however, even if a tradeoff exists, it is a priori unclear how to harness it: one would seemingly need a "tuning oracle" to adjust the procedure to yield an optimal tradeoff, by selecting optimal truncation and summarization sizes. An exhaustive search for such an adjustment is useful for illustration, but it defeats the purpose of the endeavor, which is to yield a practical algorithm whose running time decreases with more data. In this section, we address this challenge by proposing a *TRadeoff nAvigation algorithM (*TRAM*)*. It uses a limited amount of additional validation data to explore the summarization landscape, and leads to a summarization that exhibits acceptable loss in risk $\varepsilon_{\mathsf{total}}$, time $t^\star$, and space $s^\star$, thus effectively approximating a tuning oracle. We focus specifically on data-time tradeoffs via coreset data-summarization schemes, though the approach is potentially extensible to other tradeoffs and procedures.

**Theoretical Setting** We design and study our algorithm under the following assumptions.

(A) The running time of the coreset procedure is known, up to scaling. In particular, we consider a polynomial time solver and take $t_{\mathsf{core}} = \alpha m + s^\beta$ for known $\alpha, \beta > 1$.

(B) Evaluating the empirical risk using a data set of size $a$ takes a running time of $ka$.

(C) Let $m^\star$ and $s^\star$ be the solutions of Equation (3) realizing the optimal time $t^\star = t^\star_{\mathsf{core}}(n, \varepsilon_{\mathsf{total}})$. We have $R(\tilde{c}(m,s)) \leq \varepsilon_{\mathsf{total}}$ for all $m \geq m^\star$, $s \geq s^\star$, with probability at least $1 - \lambda$.

(D) We have access to additional samples from the distribution $\mathbf{P}$, beyond the data size $n$.

Assumption (A) maps to the framework of Section 5.2: $t^{\mathsf{init}}_{\mathsf{core}}(m)$ is linear in $m$, $t^{\mathsf{samp}}_{\mathsf{core}}$ is absorbed into $t_{\mathsf{solver}}$, $t_{\mathsf{solver}}$ is polynomial, and both are normalized to maintain only a single constant. Assumption (B) is trivial, except for absorbing the dimension and leading constants into $k$. (C) is a monotonicity assumption, requiring that with some probability $1 - \lambda$ not just the optimal coreset size *but also all larger* summaries are below the base risk $\varepsilon_{\mathsf{total}}$. The algorithm does not use $\lambda$, it is there only for performance analysis. Lastly, Assumption (D) uses separate data to validate in order to both use independence from the data itself, and allow to derive sample complexities for validation using basic concentration inequalities. Theorem 2 shows that only a small number of such points are needed. In practice, the data itself is partitioned to provide these points.

**A TRadeoff nAvigation algorithM** (TRAM) The idea of TRAM is as follows: search for a good summarization by starting small then growing until the desired risk is achieved. The challenge is that the risk cannot be known exactly and needs to be tested using data. We therefore have a compromise: if we stop too early we miss the target, and if we stop too late we spend too much on computation. The analysis shows that the algorithm achieves a certain balance.

**Algorithm** TRadeoff nAvigation algorithM (TRAM)

1: **Input:** Data of size $n$; risk level $\varepsilon_{\text{total}}$; validation data of size $a$; accuracy parameter $\delta > 0$.
2: **Initialization:** Start with a truncation of size $m[0] < n$ and a coreset size of $s[0]$.
3: **repeat**
4: **Iteration step** $i$**:** Summarize the $m[i]$-truncation to a coreset of size $s[i]$, and solve for the centers $\tilde{c}[i]$. Increment $m[i]$ to $m[i+1]$, and $s[i]$ to $s[i+1]$. Use a portion $a[i]$ of the validation data to evaluate the empirical risk of $\tilde{c}[i]$.
5: **until** $R_{a[i]}(\tilde{c}[i]) \leq \tau$.
6: **Output:** The last set of centers $\tilde{c}[i]$.

The validation data is a growing sequence drawn from the points described in Assumption (D). More specifically, $4b\log(1/\delta)/\varepsilon_{\text{total}}{}^2$ additional points are used at each iteration, where $b = 2B^2$, and thus $a[i] = 4ib\log(1/\delta)/\varepsilon_{\text{total}}{}^2$. The size increments happen multiplicatively: $m[i+1] \leftarrow \gamma_m m[i] \wedge n$ and $s[i+1] \leftarrow \gamma_s s[i]$. In particular, we take $\gamma_m = 2$ and $\gamma_s = 2^{1/\beta}$. Lastly, the threshold (in step 5) is $\tau = 3\varepsilon_{\text{total}}/2$.

**Theorem 2.** *Let $T$ and $J$ denote the running time and number of iterations of* TRAM *respectively. Under assumptions (A) to (D), given data of size n, a base risk $\varepsilon_{\text{total}}$, and parameter $\delta < \frac{1}{5}$, with probability at least $(1-\lambda)(1-5\delta)$,* TRAM*:*

$\triangleright$ *runs for time* $T \leq 4t^{\star 2} + \frac{8bk}{\varepsilon_{\text{total}}{}^2}\log\frac{1}{\delta}\log_2^2 t^\star$,

$\triangleright$ *uses* $a[J] \leq \frac{8b}{\varepsilon_{\text{total}}{}^2}\log\frac{1}{\delta}\log_2 t^\star$ *validation points,*

$\triangleright$ *and produces centers $\tilde{c}$ with risk $R(\tilde{c}) \leq 2\varepsilon_{\text{total}}$.*

*Proof sketch.* Using the validation test at every step, growing the set to compensate for dependencies, we control the errors of stopping too far before and too far after the optimal truncation and coreset sizes. The threshold that is slightly larger than the base risk gives us a detection margin. If these errors are not too large, the polynomial structure of the running time of the coreset procedure compounds with the geometric incrementing scheme, to lead to a computational overhead that remains reasonably close to the optimal. □

Note that the search does come with a penalty (the running time is squared). However, the analysis is very conservative and none of the constants depend on the data size $n$. Thus TRAM does indeed reproduce the qualitative behavior of the tradeoff, i.e. the running time decays as the data size increases, while the guaranteed risk remains effectively constant.

## 7 EXPERIMENTAL RESULTS

We now empirically establish the existence of tradeoffs and evaluate the performance of TRAM.

**Setup** Given a dataset $\mathscr{X} \subseteq \mathbb{R}^d$ and some $\varepsilon_{\text{total}}$, we wish to find the minimum computational cost of obtaining a $k$-means solution with risk less than or equal to $\varepsilon_{\text{total}}$. We interpret $\mathbf{P}$ as the uniform distribution over $\mathscr{X}$, hence we can compute the risk exactly. We simulate various dataset sizes by restricting individual experiments to a random subset of $\mathscr{X}$. For each pair of data size $n_i \in \mathscr{N}$ and summary size $s_j \in \mathscr{S}$ we sample $n_i$ instances i.i.d. from $\mathscr{X}$ and summarize the sample with a summary of size $s_j$ and solve the problem on the summary. We repeat the latter 50 times and report the average time and risk obtained. For the uniform subsampler, $s_j$ refers to the subsample size, and for the coresets it refers to the size of the coreset. We denote the cumulative running time of summarizing and solving the problem on the summary by $t(n_i, s_j)$ and the obtained risk by $R(n_i, s_j)$. For each procedure, let $\Lambda_{\text{proc}} = \{(n, t(n,s), R(n,s)) \mid n \in \mathscr{N}, s \in \mathscr{S}\}$.

We can now leverage $\Lambda_{\text{proc}}$ to characterize various tradeoffs. For example, to capture the data-time tradeoff for a particular size $n$ we find the minimum running time $t'$ such that $\exists(m, t', R) \in \Lambda_{\text{proc}}$, with $m < n$ and $R \leq \varepsilon_{\text{total}}$. Searching $\Lambda_{\text{proc}}$ yields Pareto-optimal boundaries of two oracles: coreset-based (ORACLE-C) and uniform-sampling-based (ORACLE-U). To show that one can navigate the space/time/data/risk tradeoffs in practice using TRAM, we showcase it alongside the oracles in Figure 2. Finding the oracles is computationally prohibitive as it entails a full grid search over $\mathscr{N}$ and $\mathscr{S}$. Nevertheless, the reported times assume the oracles simply *know* the best summarization.

**Datasets** SYNTHETIC — We generate synthetic data of 100,000 points in $\mathbb{R}^{100}$ from a mixture of Gaussians. We choose $k = 100$ centers in $[0, 100]^{100}$ and set them as means for the $k$ spherical Gaussian distributions with $\Sigma = 5I$. The relative magnitudes of the clusters are sampled from an exchangeable Dirichlet distribution with parameter $1/20$.

KDD2004BIO — The dataset of the Protein Homology Prediction Task in KDD Cup 2004, with 145,751 instances and 74 attributes that describe the match between two proteins. We fit $k$-means with $k = 150$.

CSN — The Community Seismic Network (CSN) uses smart phones with accelerometers as inexpensive seismometers for earthquake detection. Faulkner et al. [2011] compiled 7 GB of acceleration data and computed 17-dimensional feature vectors. We fit $k$-means with $k = 200$.

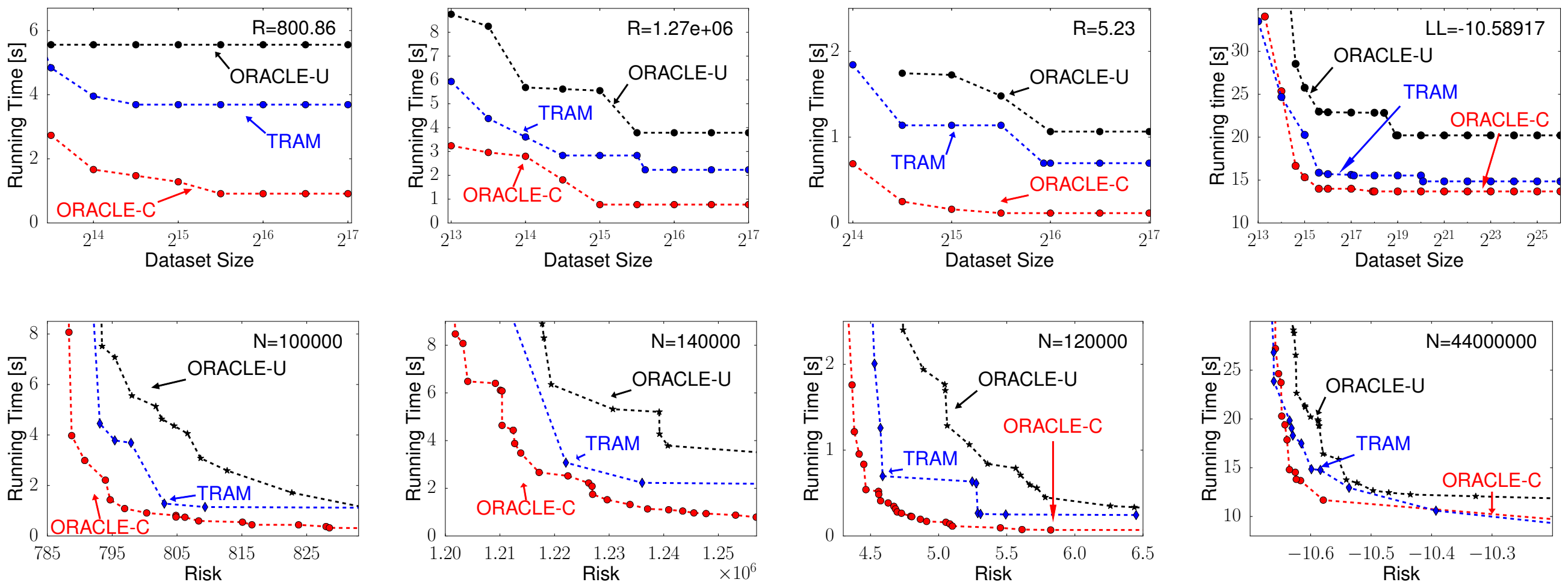

Figure 2: Results for SYNTHETIC, KDD2004BIO, CSN and WEBSCOPE datasets, per column from left to right. Figures in the first row show data-time tradeoffs: best running time for fixed risk tolerance and varying data sizes (cf. Figure 1c). Tradeoffs exist: running time decreases with increasing data size. Furthermore, the coreset procedure dominates uniform subsampling, and TRAM tracks the coreset tradeoff closely, with limited overhead. Figures in the second row show risk-time tradeoffs: best running time for fixed data size and varying risk tolerance (cf. Figure 1d).

YAHOO! WEBSCOPE R6A — $45,811,883$ instances in $\mathbb{R}^6$ that represent the user click log displayed on the Yahoo! Front Page. We fit a GMM with $k = 200$ components. The risk here is the negative log likelihood on the hold-out data.

**Parameters** For the $k$-means clustering problem we use the coreset construction from Feldman and Langberg [2011], and a weighted variant of the $k$-means++ algorithm to solve the problem on the subsample. In the case of GMMs, we use the coreset construction from Feldman et al. [2011] and a weighted EM for GMMs. We consider summarization sizes between 100 and $20,000$. For TRAM, we start with summarization size and truncation size inversely proportional to the risk required. At every iteration, we double the truncation size and take 1.5-fold ($\beta = -\log_2 1.5$) of the summarization size. $1/5^{\text{th}}$ of the data is assigned to validation, with a $\delta$ of 0.1.

**Observations** The plots in the first row in Figure 2 show the Pareto-optimal boundary for a fixed risk as data size varies. There is a data-time tradeoff as predicted from theory. Furthermore, TRAM traces the solutions achieved by the coreset oracle, implying that we *can* navigate tradeoff curves without oracles. Remarkably, TRAM remains better than the uniform subsampler oracle, eventhough either oracle takes orders of magnitude more time to obtain by exhaustive search. The second row illustrates the existence of a risk-time tradeoffs also: for fixed data size, the time to guarantee a desired risk decreases as the risk increases. Another perspective to these results is as follows. A potential practitioner is faced with three options: solving the problem on the whole dataset or doing so after summarizing, either by truncating to a portion deemed adequate or by strategically summarizing the data with a somewhat larger portion. The former is often out of the question (in the case of GMMs, it may take weeks). Summarization slashes this time down (minutes instead of weeks). However, because the coreset procedure can achieve a faster time even as it accesses a larger portion, it will be more likely to guarantee a desired risk, as compared to the uniform subsampler, at least for *interesting* (small) risk levels.

# 8 CONCLUSIONS

We explored space/time/data/risk tradeoffs achievable via coreset-based data-summarization. Our theory predicts and our empirical results demonstrate the existence and utility of such tradeoffs. We further showed how such tradeoffs can be practically realized via a novel algorithm, TRAM. While our analysis focused on $k$-means, our insights are more generally applicable. In particular, we empirically demonstrated tradeoffs in learning Gaussian Mixture Models. Approaches that optimize cost functions related to the quantization error, such as small-variance limits of non-parametric Bayesian models Jiang et al. [2012], may also immediately benefit from our results. We thus strongly believe that our results present an important step towards understanding tradeoffs in large-scale unsupervised learning. Lastly, given promising summarization-style techniques Pavlov et al. [2000], Bakir et al. [2004], Tsang et al. [2005], similar results may also be possible in supervised learning.

### Acknowledgements

This work was supported in part by ERC StG 307036, an ETH Fellowship, a Microsoft Research Faculty Fellowship, and the Zurich Information Security Center. The research was carried out when the second author was an MSR-Inria postdoctoral researcher and an ERCIM "Alain Bensoussan" fellow, funded in part by EU FP7/2007-2013 Grant 246016.

## Appendix

### Details for Proposition 2 [Modeling Error]

The proof consists of properly placing the centers. We direct the interested reader to Graf and Luschgy [2000]. Theorem 4.16 in that reference illustrates the upper bound, and the examples that follow show that the bound is tight (in $k$) in a minimax sense. We provide here an intuitive sketch to the proof, one could consider a $d$-hypercube of side $B$ rather than a sphere. We can bound the expected square-Euclidean distance with the maximal distance to a center. With a single center initially in the hypercube, this maximal distance is $dB^2/4$. We then successively split into half-sized $d$-hypercubes using cuts by $d$ $(d-1)$-hyperplanes. This multiplies the number of centers by $2^d$ and divides the distance from the centers by 2. After $\ell$ steps, we have $k = 2^{\ell d}$ centers, and the distance to the centers is bounded by $\frac{1}{2^{2\ell}} \frac{dB^2}{4}$, and thus by $\frac{dB^2}{4k^{2/d}}$.

### Details for Proposition 3 [Estimation Error]

The estimation error has been extensively studied in the statistics literature. This originated in the work of Pollard [1981], who showed that the empirically optimal centers are consistent, in the sense of both the estimation error decaying to zero and the centers converging to the optimal centers. To bound the estimation error, we use a Equation (4) of Antos et al. [2005]. Note that we can explicitly choose $\overline{\sigma} = 192$. For the lower bound we use Equation (5) of Antos et al. [2005], which holds for any estimator based on samples, and in particular for the empirically optimal centers.

### Proof of Proposition 4 [Coreset Summarization Error]

Recall Definition 1 of a $(1+\eta)$-approximation, $R_m(\tilde{c}_m) - R_m(\hat{c}_m) \leq \eta R_m(\hat{c}_m)$. Since $\hat{c}_m$ minimizes $R_m$, by taking expectations we have:

$$0 \leq \mathbf{E}[R_m(\tilde{c}_m)] - \mathbf{E}[R_m(\hat{c}_m] \leq \eta \mathbf{E}[R_m(\hat{c}_m)]. \tag{5}$$

Note that this guarantee is in terms of the empirical risk, whereas we are interested in the expected value of the true risk in the definition of the expected summarization error in Equation (2):

$$\varepsilon_{\mathsf{sum}} = |\mathbf{E}[R(\tilde{c}_m)] - \mathbf{E}[R(\hat{c}_m)]|.$$

We can relate these quantities via the following uniform concentration. By virtue of the fact that $\mathrm{d}^2(\cdot, x)$ is a smooth function on $\mathscr{C}$ (recall $\mathscr{C}$ is the set of all $k$-centers in the ball at the origin of radius $B$), as well as bounded for all $c \in \mathscr{C}$ and all $x$ in the ball, classical concentration results Linder [2002] imply that we have the following uniform concentration result, for some constant $\kappa$ (that depends on $B$ , $d$, and $k$), we have that for all $n$:

$$\mathbf{E}\left[\sup_{c\in\mathscr{C}} |R_m(c) - R(c)|\right] \leq \frac{\kappa}{\sqrt{m}}.$$

In particular, this means that for any (even random) $c$, $\mathbf{E}[|R_m(c) - R(c)|] \leq \frac{\kappa}{\sqrt{m}}$. And therefore $|\mathbf{E}[R_m(c)] - \mathbf{E}[R(c)]| \leq \frac{\kappa}{\sqrt{m}}$ as well.

$$\begin{aligned}
\mathbf{E}[R(\tilde{c})] - \mathbf{E}[R(\hat{c})] &= (\mathbf{E}[R(\tilde{c})] - \mathbf{E}[R_m(\tilde{c})]) + (\mathbf{E}[R_m(\tilde{c})] - \mathbf{E}[R_m(\hat{c})]) + (\mathbf{E}[R_m(\hat{c})] - \mathbf{E}[R(\hat{c})]) & (6)\\
&\leq \frac{2\kappa}{\sqrt{m}} + \eta \mathbf{E}[R_m(\hat{c})] \\
&\leq \frac{2\kappa}{\sqrt{m}} + \eta(\mathbf{E}[R_m(\hat{c})] - \mathbf{E}[R(\hat{c})]) + \eta \mathbf{E}[R(\hat{c})] \\
&\leq \frac{(2+\eta)\kappa}{\sqrt{m}} + \eta \mathbf{E}[R(\hat{c})], & (7)
\end{aligned}$$

where we have used the uniform concentration as well as Equation (5). Recall that $\underline{R} = \inf_{c\in\mathscr{C}} R(c)$, and that we assume $\underline{R} > 0$, that is the modeling error is positive, therefore we can write this result in a more convenient multiplicative form as follows

$$\mathbf{E}[R(\tilde{c})] - \mathbf{E}[R(\hat{c})] \leq \left(\frac{(2+\eta)\kappa}{\underline{R}\sqrt{m}} + \eta\right) \mathbf{E}[R(\hat{c})].$$

For the other side of the inequality, by using once again the fact that $\hat{c}_m$ minimizes $R_m$, we have that unlike in the above the middle term of the decomposition in Equation (6) is negative and thus:

$$\mathbf{E}[R(\hat{c})] - \mathbf{E}[R(\tilde{c})] \leq \frac{2\kappa}{\sqrt{m}}. \tag{8}$$

Since the bound in (7) is always larger than that of (8), we have:

$$|\mathbf{E}[R(\hat{c})] - \mathbf{E}[R(\tilde{c})]| \leq \left(\frac{(2+\eta)\kappa}{\underline{R}\sqrt{m}} + \eta\right)\mathbf{E}[R(\hat{c})].$$

In particular, if $\eta > \eta_0$, then by letting $m_0 = \left[\left(\frac{2}{\eta_0}+1\right)\frac{\kappa}{\underline{R}}\right]^2$ we have that for all $m > m_0$,

$$\varepsilon_{\mathsf{sum}} = |\mathbf{E}[R(\tilde{c})] - \mathbf{E}[R(\hat{c})]| \leq 2\eta\mathbf{E}[R(\hat{c})] \leq 2\eta(\varepsilon_{\mathsf{model}} + \varepsilon_{\mathsf{est}}),$$

as desired. As a remark, note how $m_0$ may depend on the parameters of the problem $(d, B, k, \underline{R})$, and the lower approximation level $\eta_0$, but not on the value of $\eta$. This would allows us to optimize over all $s_{\mathsf{core}}$ that keep $\eta(s_{\mathsf{core}})$ above $\eta_0$. □

**Proof of Theorem 1 [Existence of Tradoffs]**

In this proof, without loss of generality, we consider $t_{\mathsf{solver}}(\cdot)$, $t_{\mathsf{init}}(\cdot)$, $t_{\mathsf{samp}}(\cdot)$, $\varepsilon_{\mathsf{est}}(\cdot)$ and $\eta(\cdot)$ to be continuous functions from their domains to the real numbers. We also equate $t_{\mathsf{solver}}$ with its lower bound. To make the expressions concise, let us rename some of these functions. For the running times, let $t(x) = t_{\mathsf{solver}}(x)$, $g(x) = t_{\mathsf{core}}^{\mathsf{samp}}(x)$, and let $h$ be such that $t_{\mathsf{core}}^{\mathsf{init}}(x) = h(x)$. For the errors, let $f(x) = \varepsilon_{\mathsf{est}}(x)$, and let $\xi = \varepsilon_{\mathsf{model}}$.

The idea of the proof is as follows. Let $x$ be a given sample size and let $\varepsilon_{\mathsf{total}} = \xi + f(x)$. Since $f(x)$ is decreasing by assumption, $x$ represents the threshold of feasibility for a total risk of $\varepsilon_{\mathsf{total}}$. Note that we enunciate the statements of this proof as being for large enough $x$, which will then correspond to large enough $\varepsilon_{\mathsf{total}}$ as stated in the theorem. For all $n \geq x$, the optimal behavior of the uniform subsampler given a total risk of $\varepsilon_{\mathsf{total}}$ is to truncate to $x$ samples. Therefore, we have $t^{\star}_{\mathsf{subs}}(n, \varepsilon_{\mathsf{total}}) = t(x)$ for all $n \geq x$, and undefined otherwise. Given the coreset construction, if for some $n \geq x$ we manage to obtain a *suboptimal* coreset size $s$ that allows the coreset's total risk to match $\varepsilon_{\mathsf{total}}$ while its running time is less than $t(x)$, then the optimal running time with the optimal coreset size will only be better, implying that $t^{\star}_{\mathsf{core}}(n, \varepsilon_{\mathsf{total}}) < t^{\star}_{\mathsf{subs}}(n, \varepsilon_{\mathsf{total}})$ as claimed. Furthermore this remains true for all $n' > n$ we have $t^{\star}_{\mathsf{core}}(n', \varepsilon_{\mathsf{total}}) < t^{\star}_{\mathsf{subs}}(n', \varepsilon_{\mathsf{total}})$. Therefore the proof proceeds to show that for large enough $x$ and some $n > x$ such a choice of $s$ is possible under the conditions of the theorem.

By the fact that $t(x)$ is convex increasing by Condtion (d), for all $\alpha < 1$ there exists $\beta < 1$ (strictly below 1) such that for large enough $x$ we have $t(\alpha x) < \beta t(x)$. In particular, for $b$ as given by Condition (b), define $\delta \in (0, 1)$ to be such that for large enough $x$ we have:

$$t(x/(1+b)) < (1-2\delta)t(x). \tag{9}$$

Furthermore, we can choose $x$ large enough so that:

- By Condition (c) and the fact that, by (d), $t_{\mathsf{solver}}$ is super-linear,

$$\frac{1 - \frac{f(h^{-1}\delta t(x))}{f(x)}}{\xi + f(h^{-1}\delta t(x))} > \frac{1}{\xi} - a. \tag{10}$$

- By Condition (a),

$$g(x) \leq \delta t(x), \text{ and} \tag{11}$$

- By Condition (b),

$$2\eta(x/(1+b)) < \left(\frac{1}{\xi} - a\right) f(x). \tag{12}$$

Let $x$ be large enough so that it satisfies the above conditions and let $\varepsilon_{\text{total}} = \xi + f(x)$. Let:

$$s = t^{-1}((1-2\delta)t(x)), \tag{13}$$

and note that we have $s < x$ by virtue of $t$ being increasing. Let $m$ be the truncation size of the coreset. The coreset's total risk can therefore be written as $(\xi + f(m))(1+2\eta(s))$, based on Proposition 4. Therefore, to match a risk of $\varepsilon_{\text{total}}$, we can set:

$$m = f^{-1}\left(\frac{\xi + f(x)}{1+2\eta(s)} - \xi\right). \tag{14}$$

Choose any $n \geq m$, guaranteeing thus to have enough samples for the coreset construction. Since $f$ is increasing, note that $m \geq x$, and therefore we indeed have $n \geq x$. Since with the choice of $m$ in Equation (14) the total risk of $\varepsilon_{\text{total}}$ is satisfied, the only thing that remains to verify is that the coreset running time with these choices of $m$ and $s$ falls below the running time $t(x)$ of the uniform subsampler.

Recall that $t_{\text{core}}(m,s) = t_{\text{solver}}(s) + t^{\text{init}}_{\text{core}}(m) + t^{\text{samp}}_{\text{core}}(s)$, and we can write this explicitly as:

$$t_{\text{core}}(m,s) = t(s) + h \cdot m + g(s).$$

Our suboptimal choice of $s$ in Equation (13) is designed to carve just enough room under $t(x)$. Indeed, by construction, we have $t(s) = (1-2\delta)t(x)$. Since $s < x$ and since $g$ is increasing, we have by Equation (11) that $g(s) < g(x) < \delta t(x)$.

We are only left to verify that $h \cdot m < \delta t(x)$. By applying Equations (10), (12), (9), and (13) in that order, we have:

$$\frac{f(x) - f(h^{-1}\delta t(x))}{\xi + f(h^{-1}\delta t(x))} \overset{(10)}{>} \left(\frac{1}{\xi} - a\right) f(x) \overset{(12)}{>} 2\eta(x/(1+b)) \overset{(9)}{>} 2\eta\left(t^{-1}((1-2\delta)t(x))\right) \overset{(13)}{=} 2\eta(s)$$

It is easy to rewrite this as:

$$\frac{\xi + f(x)}{1+2\eta(s)} - \xi > f(h^{-1}\delta t(x))$$

We can now compare this to the expression for $m$ in Equation (14). In particular, using the fact that $f^{-1}$ is decreasing, we find that indeed $m < h^{-1}\delta t(x)$. Therefore, for such large enough $x$, for $n \geq m$, we have $t^{\star}_{\text{core}}(n, \varepsilon_{\text{total}}) \leq t_{\text{core}}(m,s) < (1-2\delta)t(x) + \delta t(x) + \delta t(x) = t(x)$, which concludes the proof. □

**Proof of Proposition 1 [Verification of Tradeoffs]**

We verify that Theorem 1 applies to existing coreset procedures. First, note that for a distribution for which Proposition 3 is tight, we have that $\varepsilon_{\text{est}}(x) = \Theta(1/\sqrt{x})$ (the constants do not matter here), which satisfies Condition (c). On the other hand, all generic solvers with exact guarantees have running times that increase and do so much faster than linearly, say at least polynomially $t_{\text{solver}}(x) = \Omega(x^{\beta})$ for some $\beta > 1$, and Condition (d) is easily satisfied. Furthermore, in the construction of Feldman and Langberg [2011] we saw that both $t^{\text{init}}_{\text{core}}$ and $t^{\text{samp}}_{\text{core}}$ are linear, and Condition (a) thus follows.

What remains to verify is Condition (b). Using Proposition 3, let us write more explicitly (the constants do matter here) $\varepsilon_{\text{est}}(x) \geq \underline{\sigma} B^2 \sqrt{k^{1-4/d}}/\sqrt{x}$, for large enough $x$. Consider now the Feldman and Langberg [2011] coreset, with $s_{\text{core}} = O(dk(2+\eta)^2/\eta^2)$. Note that in Feldman and Langberg [2011], this is given for $k$-median and in terms of the framework of $1 \pm \varepsilon$-coresets, written as $O(dk/\varepsilon^2)$. The result translates to $k$-means, and to the $(1+\eta)$-approximation notion of Definition 1. Write the inverse function $\eta(s_{\text{core}})$, and also interpret the bound more explicitly, as $\eta(x) < A\sqrt{dk}/(\sqrt{x} - \sqrt{dk})$, for large enough $x$ for some $A > 0$. Condition (b) is now satisfied whenever $\varepsilon_{\text{model}}(k) < \frac{\underline{\sigma}}{2\sqrt{dA}} \frac{B^2}{k^{2/d}}(1 - \sqrt{dk}/\sqrt{x})$ strictly. By comparing to Proposition 2, we see that this is satisfied for large enough $x$ whenever $A < \frac{\underline{\sigma}}{2d\sqrt{d}}$. This is a very conservative analysis, due to the looseness of the asymptotic constants, but it shows that even so, it is possible for coresets to have a leading edge on uniform subsampling. □

**Details of Numerical Simulations in Section 5.1**

Let $\varepsilon_{\text{total}}$ be a given desired risk level. The simulations consist of setting specific choices of the parameters of the problem and then, for each procedure and for each data size $n$, performing an explicit grid search over data summarization to minimize total running time while matching the $\varepsilon_{\text{total}}$ risk.

To make sure the risk is satisfied in expectation, the search is done using the upper bounds given in Section 5.2. To stay conservative for the running time, we let $t_{\mathsf{solver}}(x) = \alpha_{\mathsf{solver}} dkx^{\beta}$ with $\beta > 1$ (note that the typical dependence on $d$ and $k$ is much higher), with an explicity constant $\alpha_{\mathsf{solver}}$. We also let $t^{\mathsf{init}}_{\mathsf{core}}(m) = \alpha_{\mathsf{init}} dkm$, $t^{\mathsf{samp}}_{\mathsf{core}}(s) = \alpha_{\mathsf{samp}} s$ (recall that these are linear in their arguments), with explicit constants $\alpha_{\mathsf{init}}$ and $\alpha_{\mathsf{samp}}$. Lastly, we use the Feldman and Langberg [2011] construction for the coreset, with $\eta(s) < A\sqrt{k}/(\sqrt{s} - \sqrt{dk}) \approx A\sqrt{k}/\sqrt{s}$ (somewhat simplifying the expressions, since $dk$ is often much smaller than $s$) with an explicit constant $A$. Since $B$ appears everywhere in the errors, we can factor it out safely (set $B = 1$). We also factor out $\alpha_{\mathsf{solver}} dk$ by renormalizing the computation time (set $\alpha_{\mathsf{solver}} = 1/dk$). In summary, the remaining parameters are: $d$, $k$, $\overline{\sigma}$, $\alpha_{\mathsf{init}}$, $\alpha_{\mathsf{samp}}$, $\beta$, and $A$. Formally, we have the following.

For the uniform subsampler, the numerical version of $t^{\star}_{\mathsf{subs}}(n, \varepsilon_{\mathsf{total}})$ is:

$$
\begin{array}{ll}
\text{minimize} & m_{\mathsf{subs}}^{\beta} \\
\text{subject to} & 1 \leq m_{\mathsf{subs}} \leq n \\
& \frac{d}{k^{2/d}} + \frac{\sqrt{kd}}{\sqrt{m_{\mathsf{subs}}}} \leq \varepsilon_{\mathsf{total}}
\end{array}
$$

For the coreset procedure, the numerical version of $t^{\star}_{\mathsf{core}}(n, \varepsilon_{\mathsf{total}})$ is:

$$
\begin{array}{ll}
\text{minimize} & s_{\mathsf{core}}^{\beta} + \alpha_{\mathsf{init}} m_{\mathsf{core}} + \alpha_{\mathsf{samp}} s_{\mathsf{core}}/(dk) \\
\text{subject to} & 1 \leq m_{\mathsf{core}} \leq n \\
& 1 \leq s_{\mathsf{core}} \leq n \\
& \left(1 + 2\frac{A\sqrt{k}}{\sqrt{s_{\mathsf{core}}}}\right)\left(\frac{d}{k^{2/d}} + \frac{\sqrt{kd}}{\sqrt{m_{\mathsf{core}}}}\right) \leq \varepsilon_{\mathsf{total}},
\end{array}
$$

where if the coreset size attempts to exceed $n$, we back off to the uniform subsampler. (In the Figure, this is indicated by the fact that the coreset has no points in the initial feasile range.)

To imitate realistic values encountered in our experiments, we let the parameters be as follows: $d = 20$, $k = 20$ , $\overline{\sigma} = 192$ (the value cited in Antos et al. [2005]), $\alpha_{\mathsf{init}} = 100$, $\alpha_{\mathsf{samp}} = 100$, $\beta = 3$ and $A = 5$. We then plot the data-time tradeoff of both the subsampler (in black) and the coreset procedure (in red) in Figure 1c. We fix $\varepsilon_{\mathsf{total}} = 300$, and vary $n$.

Several observations are in order. First, the problem is not feasible for data sizes that are too small (the "data-bounded" regime). The uniform subsampler immediately throws away data upon becoming feasible, and maintains a constant computation time thereafter. If the theory is used as is, then the coreset procedure becomes more expensive for an interval of small data sizes because of lack of summarization room. To reflect what is practiced, we simply back off to the uniform subsampler when the coreset becomes too large to be cost-effective (alternatively, we may think of the entire data set as a coreset).

As more samples become available, the estimation error shrinks, which allows the coreset to pick up performance. The optimal behavior of the coreset procedure is to optimize its space summarization in order to smoothly drive the running time down (this is the "intermediate regime").

For very large data size $n$, the computation time of constructing a coreset is bound to increase. At this point, the coreset procedure opts to truncate the data just like the uniform subsampler, and therefore maintains its lowest computation time. This is why the running time of the coreset eventually flattens (this is the "data-laden" regime).

Finally, the same type of numerical optimization via grid search can be done to obtain other tradeoffs. To illustrate this, we plot the data-time tradeoff in Figure 1d, with the same parameters, for fixed $n = 2,000$ samples, and as $\varepsilon_{\mathsf{total}}$ varies.

## Proof of Theorem 2 [The TRAM Algorithm]

The proof idea is as follows. Using the validation test at every step, we control the errors of stopping too far before and too far after the optimal truncation and coreset sizes. If these errors are not too large, the polynomial structure of the running time of the coreset procedure compounds with the geometric incrementing scheme of the algorithm, to lead to a computational overhead that remains reasonably close to the optimal.

**Roadmap** We divide the proof into the following stages. We first analyze the stopping probabilities. We then characterize the number of iterations needed for feasibility. Using both of these stages, we describe the randomness of the number of total iterations. We use this to find the deviation probability of the running time within the coreset procedure as well as that of the total running time. We conclude by establishing the risk guarantees of TRAM, and bounding the amount of validation data used.

Referring to Equation (3) in Section 4, the optimal running time $t^\star_{\mathsf{core}}$ for data size $n$ and risk $\varepsilon_{\mathsf{total}}$ may be expressed in terms of an optimal truncation size $m^\star_{\mathsf{core}}$ and optimal coreset size $s^\star_{\mathsf{core}}$. By dropping the core subscript, we write $t^\star = \alpha m^\star + s^{\star\beta}$. To reduce clutter, we also write $\varepsilon_{\mathsf{total}} \equiv \varepsilon$ throughout.

We perform our analysis conditionally on the strong feasibility event of Assumption C, which we can denote by:

$$\mathscr{F} = \{R(\tilde{c}(m,s)) \leq \varepsilon_{\mathsf{total}},\ \forall m \geq m^\star, s \geq s^\star\}.$$

We bound the conditional probability $\mathbf{P}\{\mathscr{E}|\mathscr{F}\}$ of three types of events that characterize the regular behavior of TRAM:

$$\mathscr{E}_{\text{risk}} = \{R(\tilde{c}[J]) \leq 2\varepsilon\},$$

$$\mathscr{E}_{\text{time}} = \{T \leq f_T(t^\star)\},\ \text{and}$$

$$\mathscr{E}_{\text{validation}} = \{a[J] \leq f_a(t^\star)\},$$

where $J$ is the total number of iterations of TRAM, $\tilde{c}[J]$ represents its output, $T$ its total running time, $a[J]$ the number of validation data points it uses, and $f_T(\cdot)$ and $f_a(\cdot)$ are functions to be specified. For the statement of the theorem, we intersect these events (or equivalently, we union bound their complements), then we use:

$$\mathbf{P}\{\cap\mathscr{E}\} \geq \mathbf{P}\{\cap\mathscr{E}|\mathscr{F}\}\mathbf{P}\{\mathscr{F}\} \geq (1-\lambda)\mathbf{P}\{\cap\mathscr{E}|\mathscr{F}\}. \tag{15}$$

To keep the exposition clean, we drop the $|\mathscr{F}\}$ notation in what follows, implicitly understanding that all computed probabilities are conditional on $\mathscr{F}$.

**Stopping probabilities** To analyze the algorithm, we need to determine various stopping and not-stopping probabilities. For this, we use simple Hoeffding's concentration inequalities. For any fixed set of centers $c$, given $a$ sample points from the distribution $\mathbf{P}$, we have (with a constant $b = 2B^2$):

$$\mathbf{P}\{R_a(c) > R(c) + \zeta\} \leq \exp\left(-\frac{a}{b}\zeta^2\right),$$

and

$$\mathbf{P}\{R_a(c) < R(c) - \zeta\} \leq \exp\left(-\frac{a}{b}\zeta^2\right).$$

At every iteration we are attempting to test the null hypothesis $\mathscr{H}_0 = \{R(\tilde{c}) < \varepsilon\}$. We can encounter two types of errors: type I, the hypothesis holds yet we don't stop, and type II, the hypothesis does not hold yet we do stop. The threshold test helps us control type I errors primarily:

$$\begin{aligned}
\mathbf{P}\{R_{a[i]}(\tilde{c}) > \tau; \mathscr{H}_0\} &= \mathbf{P}\{R_{a[i]}(\tilde{c}) > R(\tilde{c}) + \tau - R(\tilde{c}); \mathscr{H}_0\} \\
&= \mathbf{P}\{R_{a[i]}(\tilde{c}) > R(\tilde{c}) + \varepsilon/2; \mathscr{H}_0\} \\
&\leq \mathbf{P}\left\{R_{a[i]}(\tilde{c}) > R(\tilde{c}) + \sqrt{\frac{b}{a[i]}\log\frac{1}{\delta^i}}\right\} \\
&\leq \delta^i,
\end{aligned} \tag{16}$$

where we have used Hoeffding's inequality and the fact that $a[i] = \frac{b}{(\varepsilon/2)^2}\log\frac{1}{\delta^i}$.

In the absence of an alternative hypothesis, we cannot control type II errors. A natural choice of alternative is $\mathscr{H}_1 = \{R(\tilde{c}) > 2\varepsilon\}$, which gives us the same error bound as above while incurring a factor of 2 in the potential risk. More precisely we have similarly to the above:

$$\begin{aligned}
\mathbf{P}\{R_{a[i]}(\tilde{c}) \leq \tau; \mathscr{H}_1\} &= \mathbf{P}\{R_{a[i]}(\tilde{c}) \leq R(\tilde{c}) + \tau - R(\tilde{c}); \mathscr{H}_1\} \\
&= \mathbf{P}\{R_{a[i]}(\tilde{c}) < R(\tilde{c}) - \varepsilon/2; \mathscr{H}_1\} \\
&\leq \mathbf{P}\left\{R_{a[i]}(\tilde{c}) < R(\tilde{c}) - \sqrt{\frac{b}{a[i]}\log\frac{1}{\delta^i}}\right\} \\
&\leq \delta^i
\end{aligned} \tag{17}$$

**Iterations until feasibility** Under event $\mathscr{F}$, we know that it is possible to find values of truncation $m \leq n$ and coreset size $s$ such that a risk of $\varepsilon$ is achieved and maintained for larger summarizations. The first iteration where we obtain such strong feasibility may be written as:

$$i(\varepsilon) := \min\{i : R(\tilde{c}[j]) \leq \varepsilon \; \forall j \geq i\} \tag{18}$$

Although Equation (18) is precisely the earliest we satisfy the risk condition, it makes it hard to explicitly compare to the time-optimal $m^\star$ and $s^\star$ for ultimately comparing with $t^\star$. Consider instead the following iteration:

$$i^\star = \min\{i : \gamma_m^i m[0] \geq m^\star \text{ and } \gamma_s^i s[0] \geq s^\star\} \tag{19}$$

This a conservative bound on the number of iterations until feasibility, since we know that $m^\star$ and $s^\star$ already satisfy the risk condition (recall once more that we are operating under $\mathscr{F}$), and therefore:

$$i^\star \geq i(\varepsilon).$$

Strictly speaking, this definition gives us feasibility, but it gives only an upper bound, not a lower bound on the running time. Nevertheless, since $i^\star$ (through bounding $i(\varepsilon)$) represents the main transitional point in our control of stopping conditions, we would like to express our running time relatively to it. While it is true that either one or the other of the two conditions of Equation (19) will have to fail for iterations prior to $i^\star$, we need to have both of them fail to compare to the optimal running time. It is thus convenient to define $i_m$ and $i_s$ to denote the steps just before $m[i]$ and $s[i]$ respectively reach their optimal values. That is,

$$\begin{aligned} i_m &= \lceil \log[m^\star/m[0]]/\log\gamma_m \rceil - 1, \\ i_s &= \lceil \log[s^\star/s[0]]/\log\gamma_s \rceil - 1. \end{aligned}$$

In what follows, we find it convenient to use coarse bounds:

$$i_m \leq \frac{\log m^\star}{\log\gamma_m} - 1 \text{ and } i_s \leq \frac{\log s^\star}{\log\gamma_m} - 1.$$

Using this, we can also write $i^\star = i_m \vee i_s + 1 \leq \frac{\log m^\star}{\log\gamma_m} \vee \frac{\log s^\star}{\log\gamma_m}$, therefore using $\gamma_m = 2$ and $\gamma_s = 2^{1/\beta}$ we get:

$$i^\star \leq \log_2 m^\star \vee \beta\log_2 s^\star \leq \log_2 m^\star + \beta\log_2 s^\star.$$

On the other hand, since $t^\star = \alpha m^\star + s^{\star\beta}$, we have:

$$\log t^\star \geq \frac{1}{2}\log_2 m^\star + \frac{1}{2}\beta\log_2 s^\star + \frac{1}{2}\log_2\alpha + 1,$$

where we have used the concavity of the logarithm and Jensen's inequality on $\log_2\left(\frac{1}{2}2\alpha m^\star + \frac{1}{2}2s^{\star\beta}\right)$. Combining these two observations, and noting that in our time units we have $\alpha \geq 1$ and thus $\log_2\alpha \geq 0$, we have:

$$i^\star \leq 2\log_2 t^\star - 2. \tag{20}$$

**Total number of iterations** Recall that $J$ denotes the number of iterations of the algorithm. We can split the iterations into two halves, before and after the problem becomes feasible. In light of the stopping probabilities that we analyzed, we can therefore determine the behavior of $J$.

**Proposition 5.** *If $i \geq i^\star$ then:*

$$\mathbf{P}\{J > i\} \leq \mathbf{P}\{J > i | J \geq i^\star\} \leq \delta^i.$$

*Proof.* We can write $\mathbf{P}\{J > i\} = \mathbf{P}\{J > i | J < i^\star\}\mathbf{P}\{J < i^\star\} + \mathbf{P}\{J > i | J \geq i^\star\}\mathbf{P}\{J \geq i^\star\}$. If $i \geq i^\star$, the first product is null, and the second is no greater than the $\mathbf{P}\{J > i | J \geq i^\star\}$ term.

As the validation data is independent from the samples used to generate $\tilde{c}[j]$ and since we are operating under event $\mathscr{F}$, we have that for all iterations $j \geq i^\star$ hypothesis $\mathscr{H}_0$ holds and from Equation (16):

$$\mathbf{P}\{\text{TRAM does not stop at } j\} = \mathbf{P}\{R_{a[j]}(\tilde{c}[j]) > \tau\} \leq \delta^j.$$

Using this, we can do an intersection bound for all $i \geq i^\star$:

$$\begin{aligned}\mathbf{P}\{J > i | J \geq i^\star\} &= \mathbf{P} \bigcap_{j=i^\star}^{i} \{\text{TRAM does not stop at } j\} \\ &\leq \min_{j=i^\star}^{i} \mathbf{P}\{\text{TRAM does not stop at } j\} \\ &\leq \delta^i,\end{aligned}$$

which establishes the claim. □

**Total running time** Let $T_1$ denote the total running time of the coreset procedure computations. Let $T_2$ denote the running time of the validation computations. The total computation can then be written as the sum of these two running times $T = T_1 + T_2$. As we expect, the lion's share of the computation is consumed by the coreset procedure. We now analyze $T_1$ and $T_2$ separately.

Let us first write out the explicit dependence of $T_1$ on $J$:

$$T_1[J] := T_1 = \sum_{i=0}^{J} \left(\alpha m[i] + s[i]^\beta\right) = \sum_{i=0}^{J} 2^i \left(\alpha m[0] + s[0]^\beta\right)$$

where for the last expression recall that we use $\gamma_m = 2$ and $\gamma_s = 2^{1/\beta}$ for the multiplicative increment at each iteration: $m[i+1] \leftarrow \gamma_m m[i] \wedge n$ and $s[i+1] \leftarrow \gamma_s s[i]$.

Our goal is to bound the probability of the event that $T_1$ is well behaved, and in particular to show that with high probability $T_1 \leq (1+v)2t^{\star 2}$ for arbitrary $v \geq 1$. To do so, we fist show that the running time until (but excluding) $i^\star$ is bounded by $2t^{\star 2}$. We then show that, thanks to Proposition 5, the likelihood of stopping is so high beyond $i^\star$ that we only incur the additional $(1+v)$ factor, with high probability. We can define the running time before $i^\star$ and bound it as follows:

$$T_1[i^\star - 1] := \sum_{i=0}^{i^\star - 1} \left(\alpha m[i] + s[i]^\beta\right) = \sum_{i=0}^{i^\star - 1} 2^{i - i^\star + 1} \left(\alpha m[i^\star - 1] + s[i^\star - 1]^\beta\right) \leq 2\left(\alpha m[i^\star - 1] + s[i^\star - 1]^\beta\right) \tag{21}$$

Based on this, to compare the running time $T_1[i^\star - 1]$ to the optimal running time $t^\star = \alpha m^\star + s^{\star\beta}$, it suffices to compare $\alpha m[i^\star - 1] + s[i^\star - 1]$ to this optimum. For the moment, in order to have a clean book-keeping, let us revert to the symbols $\gamma_m$ and $\gamma_s$ of the multiplicative updates, instead of their numerical values.

In particular, the following peculiarity will occur: either $m^\star$ will be reached first or $s^\star$ will. Let us treat each of the following two cases separately:

- Say $m^\star$ is reached first, that is $i_m \leq i_s$. In this case, $i^\star = i_s + 1$, and we have $m[i_s] < \gamma_m{}^{\log s^\star / \log \gamma_s} m^\star$. The factor in the latter expression uses the fact that $m[i_m] < m^\star$ and that $m$ is continued to be incremented for at most $i_s - i_m + 1 \leq i_s + 1 \leq \log s^\star / \log \gamma_s$ steps, by a multiple of $\gamma_m$ at each step. Noting also that $s[i_s] < s^\star$, we have:

$$\begin{aligned}\alpha m[i^\star - 1] + s^\beta[i^\star - 1] &< \alpha \gamma_m{}^{\log s^\star / \log \gamma_s} m^\star + s^{\star\beta} \\ &= \alpha s^{\star \log \gamma_m / \log \gamma_s} m^\star + s^{\star\beta}.\end{aligned}$$

- Say $s^\star$ is reached first, that is $i_s \leq i_m$, parallel calculations give us that:

$$\begin{aligned}\alpha m[i^\star - 1] + s^\beta[i^\star - 1] &< \alpha m^\star + \gamma_s{}^{\beta \log m^\star / \log \gamma_m} s^{\star\beta} \\ &= \alpha m^\star + m^{\star \beta \log \gamma_s / \log \gamma_m} s^{\star\beta}.\end{aligned}$$

With our choices of $\gamma_m$ and $\gamma_s$ we have $\log \gamma_m / \log \gamma_s = \beta$, therefore in either case:

$$\begin{aligned}\alpha m[i^\star - 1] + s[i^\star - 1]^\beta &< \left(\alpha s^{\star\beta} m^\star + s^{\star\beta}\right) \vee \left(\alpha m^\star + m^\star s^{\star\beta}\right) \\ &< t^{\star 2}, \end{aligned} \tag{22}$$

where we have again used the fact that in our time units we have $\alpha \geq 1$. Of course, this bound is overly conservative. For one, we are factoring in extra increments on both terms, whereas only holds. Also, neither $m$ nor $s$ can be incremented arbitrarily many times, as they are bounded by $n$. Despite this conservatism, we adhere to this expression for its simplicity, and use it to bound the randomness of $T_1$. Going back to Equation (21) and using Equation (22), we have:

$$T_1[i^\star - 1] \leq 2\left(\alpha m[i^\star - 1] + s[i^\star - 1]^\beta\right) \leq 2t^{\star 2},$$

Choose some $v \geq 0$. From this bound, we can see that if $J < i^\star$ we will always have $T_1[J] \leq (1+v)2t^{\star 2}$, and thus $\mathbf{P}\{T_1 > (1+v)2t^{\star 2} | J < i^\star\} = 0$. Therefore, by total probability, it follows that $\mathbf{P}\{T_1 > (1+v)2t^{\star 2}\} \leq \mathbf{P}\{T_1 > (1+v)2t^{\star 2} | J \geq i^\star\}$ (cf. the proof of Proposition 5). Next when $J \geq i^\star$, we have on one hand that $T_1 > (1+v)2t^{\star 2}$ implies that:

$$\sum_{i=i^\star}^{J} \left(\alpha m[i] + s[i]^\beta\right) > (1+v)2t^{\star 2} - T_1[i^\star - 1] > 2vt^{\star 2}.$$

On the other hand, using the bound in Equation (22), we have:

$$\sum_{i=i^\star}^{J} \left(\alpha m[i] + s[i]^\beta\right) \leq \sum_{i=i^\star}^{J} 2^{i-i^\star+1} t^{\star 2} = 2\left(2^{J-i^\star+1} - 1\right) t^{\star 2}.$$

Therefore when $J \geq i^\star$, $T_1 > (1+v)2t^{\star 2}$ implies that $2^{J-i^\star+1} - 1 > v$, and thus $J > \log_2(1+v) - 1 + i^\star$. Starting with this chain of implications to bound the probabilities of their respective events, and then combining with Proposition 5, we have:

$$\begin{aligned}
\mathbf{P}\{T_1 > (1+v)2t^{\star 2}\} &\leq \mathbf{P}\{T_1 > (1+v)2t^{\star 2} | J \geq i^\star\} \\
&\leq \mathbf{P}\{J > \log_2(1+v) - 1 + i^\star | J \geq i^\star\} \\
&\leq \delta^{\log_2(1+v)-1+i^\star}
\end{aligned} \tag{23}$$

Now let us move to $T_2$, which recall denotes the running time of the validations. Per Assumption (B), we can write it explicitly as a function of the number of iterations:

$$T_2[J] = \sum_{i=1}^{J} ka[i].$$

This is easy to account for, since:

$$a[i] = i \tfrac{4b}{\varepsilon^2} \log \tfrac{1}{\delta},$$

and thus

$$T_2[J] = J(J+1) \underbrace{\tfrac{2bk}{\varepsilon^2} \log \tfrac{1}{\delta}}_{:=\gamma}.$$

We can now use arguments similar to the analysis for $T_1$ above. A useful observation in this regard is to use Equation (20) to get $(2\log_2 t^\star + w)^2 \geq (i^\star + w)(i^\star + w + 1)$, for any given $w \geq 0$. Then, by Proposition 5:

$$\begin{aligned}
\mathbf{P}\left\{T_2 > \gamma(2\log_2 t^\star + w)^2\right\} &\leq \mathbf{P}\{T_2 > \gamma(i^\star + w)(i^\star + w + 1)\} \\
&= \mathbf{P}\{\gamma J(J+1) > \gamma(i^\star + w)(i^\star + w + 1)\} \\
&= \mathbf{P}\{J > i^\star + w\} \\
&\leq \delta^{w+i^\star}
\end{aligned}$$

By restricting $v \geq 1$ and choosing $w = \log_2(1+v) - 1$, we obtain a comparable exponent to $T_1$:

$$\mathbf{P}\left\{T_2 > \gamma \log_2^2\left(\tfrac{1}{2}(1+v)t^{\star 2}\right)\right\} \leq \delta^{\log_2(1+v)-1+i^\star} \tag{24}$$

By using a union bound to combine Equations (23) and (24), we have:

$$\mathbf{P}\left\{T = T_1 + T_2 > (1+v)2t^{\star 2} + \log_2^2\left(\tfrac{1}{2}(1+v)t^{\star 2}\right) \tfrac{2bk}{\varepsilon^2} \log \tfrac{1}{\delta}\right\} \leq 2\delta^{\log_2(1+v)-1+i^\star} \leq 2\delta^{i^\star} \leq 2\delta, \tag{25}$$

where we have simplified from the general form for $v \geq 1$ and using the fact that $i^\star \geq 1$.

**Risk guarantees** We would now like to use Equation (17) to say something about performance guarantees for the output of the algorithm. The probability of the stopping condition being satisfied at a given iteration $j$ when the risk is larger than $2\varepsilon$ (denote this event by $E_j$) is bounded by Equation (17):

$$\mathbf{P}\{E_j\} = \mathbf{P}\{R_{a[j]}(\tilde{c}[j]) \leq \tau \text{ and } R(\tilde{c}[j]) > 2\varepsilon\} \leq P\{R_{a[j]}(\tilde{c}[j]) \leq \tau | R(\tilde{c}[j]) > 2\varepsilon\} \leq \delta^j,$$

where we have used the fact that the validation data is independent from the samples, and that the bound of Equation (17) holds for any center $\tilde{c}$ such that $R(\tilde{c}) > 2\varepsilon$.

Therefore, the conditional probability of actually stopping at any such $j$ is given by:

$$\mathbf{P}\{R(\tilde{c}[J]) > 2\varepsilon\} \leq \mathbf{P}\bigcup_j E_j \leq \sum_{j=1}^{\infty} \delta^j \leq \frac{\delta}{1-\delta} \leq 2\delta, \tag{26}$$

since $\delta < \frac{1}{5} < \frac{1}{2}$. Therefore, no matter how many iterations we have, the probability that we will stop at a high-risk iteration is bounded. Namely, with probability at least $1 - 2\delta$ we guarantee a risk of at most $2\varepsilon$.

**Validation data requirements** We conclude with showing that the number of validation data points needed remains small. Since the same points are reused from one iteration to the other, the total amount of validation data used is:

$$a = a[J] = J\tfrac{4b}{\varepsilon^2}\log\tfrac{1}{\delta}.$$

By Proposition 5 and Equation (20) we have that:

$$\mathbf{P}\{a[J] > 2\log_2 t^\star \tfrac{4b}{\varepsilon^2}\log\tfrac{1}{\delta}\} \leq \mathbf{P}\{a[J] > i^\star \tfrac{4b}{\varepsilon^2}\log\tfrac{1}{\delta}\} \leq \mathbf{P}\{J > i^\star\} \leq \delta^{i^\star} \leq \delta. \tag{27}$$

**Synthesis** The theorem follows by union bounding (25) (with $v = 1$), (26), and (27) and using (15). □